\documentclass[runningheads]{llncs}
\usepackage[T1]{fontenc}
\usepackage[inline]{enumitem}
\usepackage{makecell}
\usepackage{graphicx}
\begin{document}
\title{BERT(s) to Detect Multiword Expressions}
%
%

\author{Damith Premasiri \and
 Tharindu Ranasinghe}
\authorrunning{D. Premasiri et al.}
%
\institute{University of Wolverhampton, UK
\email{\{damith.premasiri,tharindu.ranasinghe\}@wlv.ac.uk}}
\maketitle              
\begin{abstract}
Multiword expressions (MWEs) present groups of words in which the meaning of the whole is not derived from the meaning of its parts. The task of processing MWEs is crucial in many natural language processing (NLP) applications, including machine translation and terminology extraction. Therefore, detecting MWEs is a popular research theme. In this paper, we explore state-of-the-art neural transformers in the task of detecting MWEs. We empirically evaluate several transformer models in the dataset for SemEval-2016 Task 10: Detecting Minimal Semantic Units and their Meanings (DiMSUM). We show that transformer models outperform the previous neural models based on long short-term memory (LSTM). The code and pre-trained model will be made freely available to the community. 

\keywords{Multiword Expressions  \and Transformers \and Deep Learning.}
\end{abstract}
\section{Introduction}
The term “multiword expressions” (MWEs) denotes a group of words that act as a morphologic, syntactic and semantic unit in linguistic analysis; however, their meaning cannot be inferred from the meaning of their components \cite{boros-etal-2017-data}. For example, the MWE \textit{"by and large"} have a meaning equivalent to \textit{"on the whole"}. But none of the words in the MWE imply this \cite{10.3115/1119282.1119294}. MWEs can be categorised in to different categories such as lexicalised phrases and institutionalised phrases; however the basic definition remains same in all the categories. MWEs appear in almost all languages and is a common method of expressing ideas.

Apart from the difficulty of deriving meaning from individual components, which is known as non-compositionality in phraseology, MWEs have several challenges when processing them computationally \cite{constant-etal-2017-survey}. \begin {enumerate*} \item MWEs are non-substitutable, which means that the components of MWE cannot be replaced by synonyms (\textit{e.g., by and big}). \item MWEs and non-MWEs can be ambiguous (\textit{e.g., by and large, we agree vs he walked by and large tractors passed him}) \end{enumerate*}. These unique challenges in MWEs raise several fundamental problems with many NLP applications. For example, parsing and machine translation (MT) \cite{mitkov2018multiword,mitkov2016computational}, which depends on a clear distinction between word tokens and phrases, has to be re-thought to accommodate MWEs \cite{constant-etal-2017-survey,taslimipoor2015using}. The usual approach in these applications is to identify MWEs first, and then treat them accordingly. Therefore, detecting MWEs is a key research area in NLP.  

In recent years, the identification of MWEs has been modelled as a supervised machine learning task where the machine learning models are trained on an annotated dataset. As we explain in Section \ref{sec:related_work}, several datasets have been released to train these machine learning models. Furthermore shared tasks such as SemEval-2016 Task 10 \cite{schneider-etal-2016-semeval} and PARSEME \cite{savary-etal-2017-parseme} have contributed to develop datasets. In recent years, neural network-based models, and in particular architectures incorporating Recurrent
Neural Networks (RNNs) such as Long Short Term Memory (LSTM) and Convolutional Neural Networks (CNNs) have achieved state-of-the-art performance in MWE identification tasks \cite{savary-etal-2017-parseme}. Usually, these models utilise pre-trained word embedding models such as word2vec \cite{NIPS2013_9aa42b31} and glove \cite{pennington-etal-2014-glove}. We describe these models in Section \ref{sec:related_work}. However, these traditional word embeddings provide the same embedding for polysemous  words \cite{pathirana2019concept} \cite{pathirana2018knowledge}. Therefore, non-substitutability and the ambiguous nature of the MWEs can cause complications with traditional word embeddings. 

A possible solution is to utilise neural architectures such as transformers that incorporate context more into the learning process. However, as far as we know, there has not been any research done to compare the performance of different transformer models in the MWE identification task. In this research, we empirically evaluate several transformer models in detecting MWEs to fill this gap. The findings of this research can be beneficial for many NLP applications that require detecting MWEs.

The main contributions of this study are, 

\begin{enumerate}
    \item We empirically evaluate eight different transformer models in the task of detecting multiword expressions using a recent dataset released for SemEval-2016 Task 10 \cite{schneider-etal-2016-semeval}. 
    \item We show that transformer-based models to identify multiword expressions outperform previous neural models based on LSTMs. 
    \item We provide important resources to the community: the code as an open-source framework, as well as the pre-trained models will be freely available to the community on HuggingFace \cite{wolf-etal-2020-transformers} model hub\footnote{The public GitHub repository is available on \url{https://github.com/DamithDR/MultiwordExpressions} and the pre-trained models are available on \url{https://huggingface.co/Damith/mwe-xlm-roberta-base}}. Furthermore, we have created a docker image of the experiments adhering to the ACL reproducibility criteria\footnote{The docker image is available on \url{https://hub.docker.com/r/damithpremasiri/transformer-based-mwe}}. 
    
\end{enumerate}

\section{Related Work}
\label{sec:related_work}
As mentioned before, a clear majority of the recent research to detect MWEs are neural based models. Usually the MWE detection is modelled as a token classification task where the model predicts whether a certain token belongs to a MWE or not. Therefore this task is similar to a named entity recognition (NER) task and the models that were used for NER has been used in MWE detection too \cite{rohanian-etal-2019-bridging}. 

The most popular method to detect MWEs are based on recurrent neural network variants such as LSTMs and gated recurrent units (GRUs) \cite{rohanian-etal-2019-bridging}. \cite{moreau-etal-2018-crf} use a LSTM model with Conditional random field (CRF) to detect MWEs. Furthermore, they incorporate  dependency parse information to improve the results. Graph convolutional neural networks (GCNs) \cite{kipf2016semi} have also been applied to MWE identification. \cite{rohanian-etal-2019-bridging} incorporate multi-head self-attention to improve the performance of GCN in MWE detection. Transformers have also been used to detect MWEs\cite{chakraborty2020identification},\cite{kanclerz2022deep}; however, the research has been limited to a few transformer models. Therefore, in this research, we fill this gap by empirically evaluating multiple transformers in the task of MWE identification.  

\section{Data}
The dataset we used was from the 2016 SemEval shared task 10\footnote{SemEval 2016 shared task description: \url{http://dimsum16.github.io/}} \cite{schneider-etal-2016-semeval}. The shared task was designed to predict both minimal semantic units and semantic classes (supersenses). The training data combines and harmonises three data-sets, the STREUSLE 2.1\footnote{The STREUSLE 2.1 is available on : \url{http://www.cs.cmu.edu/~ark/LexSem/}} corpus of web reviews, as well as the Ritter and Lowlands Twitter datasets\footnote{Twitter dataset is available on : \url{https://github.com/coastalcph/supersense-data-twitter}}. The Ritter and Lowlands datasets have been reannotated for MWEs and supersenses to improve their quality and to more closely follow the conventions used in the STREUSLE annotations. The DiMSUM data files have tab-separated columns in the spirit of CoNLL, with blank lines to separate sentences. Each row contained nine columns : token offset, word, lowercase lemma, POS, MWE tag, offset of parent token (i.e. previous token in the same MWE), strength level encoded in the tag, supersense label and sentence ID. In this research we used only the word and the MWE tag. There are multiple MWE tagging formats such as IOB and IOB2. The dataset contains the IOB format where I - Inside, O - Outside, B - Beginning of a MWE. The I- prefix indicates that the tag is inside a chunk. An O indicates that a token belongs to no chunk. The B- prefix indicates that the tag is the beginning of a chunk that immediately follows another chunk without O tags between them. 

The data composition is shown in the table \ref{tab:data_table}. In the initial test dataset, there were 16500 words with 1000 sentences, however we had to remove one sentence from the test set due to encoding issues faced with the Python libraries. 

\begin{table}[ht]
    \centering
    \resizebox{7cm}{!}{
        \begin{tabular}{c|c|c}
            \hline
            Dataset & No of Words & No of Sentences  \\
            \hline
            Train & 73826 & 4800 \\
            Test & 16400 & 999
        \end{tabular}
    }
    \caption{Datasets composition}
    \label{tab:data_table}
\end{table}

\section{Methodology}
The main motivation behind the methodology is the state-of-art results produced by transformers in multiple different NLP tasks such as question answering \cite{damith2022DTWquranqa}, machine translation quality estimation \cite{ranasinghe-etal-2021-exploratory}, cyber bullying \cite{morgan-etal-2021-wlv} \cite{sarkar-etal-2021-fbert-neural}, language identification \cite{jauhiainen-etal-2021-comparing} and named entity recognition\cite{arkhipov-etal-2019-tuning}. We experiment with two types of models, which we explain in the following sections. 

\begin{figure}[ht]
\centering
\includegraphics[scale=0.5]{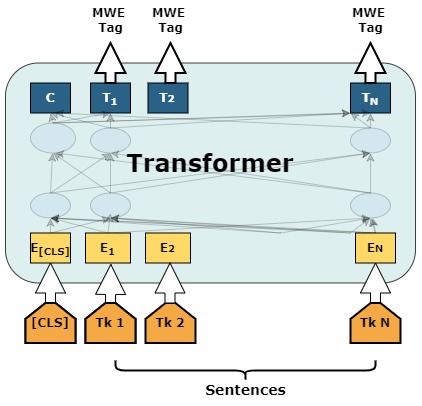}
\caption{Transformer Architecture for MWE classifier}
\label{fig:architecture}
\end{figure}

\paragraph{Transformer} models such as BERT \cite{devlin-etal-2019-bert} have been trained using masked language modelling objective and then can be fine-tuned for multiple different tasks \cite{alloatti-etal-2019-real}. This research uses the pre-trained transformer models for a token classification task. As shown in Figure \ref{fig:architecture}, we added a token level classifier on top of the transformer
model. The token-level classifier is a linear layer that takes the last hidden state of the sequence as the input and produces a label for each token as the output. In this case, each token
can have three labels; B, I and O.


\noindent We experimented with several popular, widely used transformer models to detect MWEs. Namely they are BERT \cite{devlin-etal-2019-bert}, RoBERTa \cite{zhuang-etal-2021-robustly}, XLNet \cite{NEURIPS2019_dc6a7e65}, XLM-RoBERTa \cite{conneau-etal-2020-unsupervised} and Electra \cite{clark2020electra}. For BERT we used several variations such as bert-base-cased, bert-base-uncased, bert-base-multilingual-cased and bert-base-multilingual-uncased while for other transformer model we only used the available base model. All the transformer-based methods were experimented using batch size 32, Adam optimiser with learning rate 4e-5. They were trained for 3 epochs with linear learning rate warm-up over 10\% of the training data. These experiments were done in an NVIDIA GeForce RTX 2070 GPU.



\paragraph{BiLSTM-CRF} is another token classification architecture which provided state-of-the-art results before transformers \cite{Huang2015BidirectionalLM} . Bidirectional LSTM (BiLSTM) is capable of learning contextual information both forwards and backwards in time
compared to conventional LSTMs. In this study, we used the Bi-LSTM architecture given this bidirectional ability to model temporal dependencies. CRFs \cite{10.5555/645530.655813} are a statistical model that are capable of incorporating context information and are highly
used for sequence labeling tasks. A CRF connected
to the top of the Bi-LSTM model provides a powerful way to model relationships between consecutive outputs (across time) and provides a means to efficiently utilise past and future tag information to predict the current tag of word. For the experiments, we used a learning rate 1.5e-1 and the model was trained for 50 epochs. BiLSTM-CRF experiments were conducted on a CPU.

\section{Results}
In this section, we report and compare the results of our experiments using standard evaluation metrics Weighted Recall, Weighted Precision, Weighted F1 and Macro F1 for the MWEs detection task. As shown in the Table \ref{tab:results_table} it is clear that the transformer-based models outperform the BiLSTM-CRF method with clear margins. The BiLSTM-CRF could achieve only 0.8253 and 0.3135 for Weighted F1 and Macro F1 scores, respectively, while all the transformer models we experimented outperform that. A clear observation is that even though BiLSTM-CRF has a fairly high Weighted F1 score, the Macro F1 score is very low. Since the Macro F1 score is sensitive to class imbalance, we hypothesise that this model is struggling to predict some specific label(s). On the other hand, transformer models achieve a high Macro F1 score suggesting that they can predict all the classes equally.

\begin{table}[ht]
    \centering
    \resizebox{12cm}{!}{
        \begin{tabular}{c|c|c|c|c}
            \bf Model & \bf \makecell{Weighted \\ Recall} & \bf \makecell{Weighted \\ Precision} & \bf \makecell{Weighted \\ F1} & \bf \makecell{Macro \\ F1} \\
            \hline
            roberta-base & 0.9125 & 0.9087 & 0.9103 & 0.7170 \\
            xlm-roberta-base & 0.9194 & 0.9152 & \textbf{0.9169} & \textbf{0.7366} \\
            xlnet-base-cased & 0.9179 & 0.9135 & 0.9152 & 0.7317 \\
            bert-base-multilingual-cased & 0.9086 & 0.9042 & 0.9061 & 0.7049 \\
            bert-base-multilingual-uncased & 0.9056 & 0.8990 & 0.9016 & 0.6863 \\
            bert-base-cased & 0.9169 & 0.9122 & 0.9140 & 0.7290 \\
            bert-base-uncased & 0.9087 & 0.9026 & 0.9050 & 0.6975 \\
            electra-base-discriminator & 0.9125 & 0.9062 & 0.9085 & 0.7071 \\
            \hline
            BiLSTM-CRF & 0.8807 & 0.8059 & 0.8253 & 0.3135 \\
        \end{tabular}
    }
    \caption{Results for different methods for multiword expression detection}
    \label{tab:results_table}
\end{table}

Results of transformers based neural methods have similar performance with slight differences from one model to another. It is clear that the best performer is the xlm-roberta-base model, which could achieve the best performance for both Weighted F1 and Macro F1 over all other models by achieving scores of 0.9169, 07366 accordingly. This is followed by the xlnet-base-cased model with a Macro F1 score of 0.7317, showing the competitiveness of the transformer models in MWE detection tasks. Interestingly, a multilingual model such as xlm-roberta-base could outperform language-specific transformer models on MWEs detecting task on this dataset. 

Another interesting observation is that the cased models outperform the uncased models. This is similar in both bert and bert-multilingual models, where the cased models slightly outperform the uncased models. We believe that cased models can perform better in detecting MWEs than uncased models according to this dataset.

Overall, transformers based neural methods perform higher than BiLSTM-CRF. It is clear that the results of all the transformer-based methods varied between 0.6863 - 0.7366 of Macro F1, showing their strong and competitive performance in MWE detection tasks.

\section{Conclusion}
MWE detection is an important research area for many NLP applications. In this paper, we empirically evaluate several neural transformer models in the MWE detection task using a recent dataset released for SemEval-
2016 Task 10 and show that all the transformer models outperform the LSTM based method. From the experimented transformer models, xlm-roberta-base provided the best results outperforming other transformer models. We can conclude that transformer models can handle the challenges presented by MWEs better than the previous LSTM based methods.

In the future, we would like to explore the cross-lingual capabilities of the transformer models in the MWE detection task. Cross-lingual transformer models such as xlm-roberta can be used to transfer knowledge between languages so that a model can be trained only on English data but can be used to predict on other languages. Since the xlm-roberta-based performed best in this study, we believe that this model can be further explored to detect MWEs in different languages. 
%
%
%
\bibliographystyle{splncs04}
\bibliography{mybibliography}

\begin{thebibliography}{10}
\providecommand{\url}[1]{\texttt{#1}}
\providecommand{\urlprefix}{URL }
\providecommand{\doi}[1]{https://doi.org/#1}

\bibitem{alloatti-etal-2019-real}
Alloatti, F., Di~Caro, L., Sportelli, G.: Real life application of a question
  answering system using {BERT} language model. In: Proceedings of the 20th
  Annual SIGdial Meeting on Discourse and Dialogue. pp. 250--253. Association
  for Computational Linguistics, Stockholm, Sweden (Sep 2019).
  \doi{10.18653/v1/W19-5930}, \url{https://aclanthology.org/W19-5930}

\bibitem{arkhipov-etal-2019-tuning}
Arkhipov, M., Trofimova, M., Kuratov, Y., Sorokin, A.: Tuning multilingual
  transformers for language-specific named entity recognition. In: Proceedings
  of the 7th Workshop on Balto-Slavic Natural Language Processing. pp. 89--93.
  Association for Computational Linguistics, Florence, Italy (Aug 2019).
  \doi{10.18653/v1/W19-3712}, \url{https://aclanthology.org/W19-3712}

\bibitem{10.3115/1119282.1119294}
Baldwin, T., Bannard, C., Tanaka, T., Widdows, D.: An empirical model of
  multiword expression decomposability. In: Proceedings of the ACL 2003
  Workshop on Multiword Expressions: Analysis, Acquisition and Treatment -
  Volume 18. p. 89–96. MWE '03, Association for Computational Linguistics,
  USA (2003). \doi{10.3115/1119282.1119294},
  \url{https://doi.org/10.3115/1119282.1119294}

\bibitem{boros-etal-2017-data}
Boros, T., Pipa, S., Barbu~Mititelu, V., Tufis, D.: A data-driven approach to
  verbal multiword expression detection. {PARSEME} shared task system
  description paper. In: Proceedings of the 13th Workshop on Multiword
  Expressions ({MWE} 2017). pp. 121--126. Association for Computational
  Linguistics, Valencia, Spain (Apr 2017). \doi{10.18653/v1/W17-1716},
  \url{https://aclanthology.org/W17-1716}

\bibitem{chakraborty2020identification}
Chakraborty, S., Cougias, D., Piliero, S.: Identification of multiword
  expressions using transformers  (2020)

\bibitem{clark2020electra}
Clark, K., Luong, M.T., Le, Q.V., Manning, C.D.: {ELECTRA}: Pre-training text
  encoders as discriminators rather than generators. In: ICLR (2020),
  \url{https://openreview.net/pdf?id=r1xMH1BtvB}

\bibitem{conneau-etal-2020-unsupervised}
Conneau, A., Khandelwal, K., Goyal, N., Chaudhary, V., Wenzek, G., Guzm{\'a}n,
  F., Grave, E., Ott, M., Zettlemoyer, L., Stoyanov, V.: Unsupervised
  cross-lingual representation learning at scale. In: Proceedings of the 58th
  Annual Meeting of the Association for Computational Linguistics. pp.
  8440--8451. Association for Computational Linguistics, Online (Jul 2020).
  \doi{10.18653/v1/2020.acl-main.747},
  \url{https://aclanthology.org/2020.acl-main.747}

\bibitem{constant-etal-2017-survey}
Constant, M., Eryi{\v{g}}it, G., Monti, J., van~der Plas, L., Ramisch, C.,
  Rosner, M., Todirascu, A.: {S}urvey: Multiword expression processing: A
  {S}urvey. Computational Linguistics  \textbf{43}(4),  837--892 (Dec 2017).
  \doi{10.1162/COLI_a_00302}, \url{https://aclanthology.org/J17-4005}

\bibitem{devlin-etal-2019-bert}
Devlin, J., Chang, M.W., Lee, K., Toutanova, K.: {BERT}: Pre-training of deep
  bidirectional transformers for language understanding. In: Proceedings of the
  2019 Conference of the North {A}merican Chapter of the Association for
  Computational Linguistics: Human Language Technologies, Volume 1 (Long and
  Short Papers). pp. 4171--4186. Association for Computational Linguistics,
  Minneapolis, Minnesota (Jun 2019). \doi{10.18653/v1/N19-1423},
  \url{https://aclanthology.org/N19-1423}

\bibitem{Huang2015BidirectionalLM}
Huang, Z., Xu, W., Yu, K.: Bidirectional lstm-crf models for sequence tagging.
  ArXiv  \textbf{abs/1508.01991} (2015)

\bibitem{jauhiainen-etal-2021-comparing}
Jauhiainen, T., Ranasinghe, T., Zampieri, M.: Comparing approaches to
  {D}ravidian language identification. In: Proceedings of the Eighth Workshop
  on NLP for Similar Languages, Varieties and Dialects. pp. 120--127.
  Association for Computational Linguistics, Kiyv, Ukraine (Apr 2021),
  \url{https://aclanthology.org/2021.vardial-1.14}

\bibitem{kanclerz2022deep}
Kanclerz, K., Piasecki, M.: Deep neural representations for multiword
  expressions detection. In: Proceedings of the 60th Annual Meeting of the
  Association for Computational Linguistics: Student Research Workshop. pp.
  444--453 (2022)

\bibitem{kipf2016semi}
Kipf, T.N., Welling, M.: Semi-supervised classification with graph
  convolutional networks. arXiv preprint arXiv:1609.02907  (2016)

\bibitem{10.5555/645530.655813}
Lafferty, J.D., McCallum, A., Pereira, F.C.N.: Conditional random fields:
  Probabilistic models for segmenting and labeling sequence data. In:
  Proceedings of the Eighteenth International Conference on Machine Learning.
  p. 282–289. ICML '01, Morgan Kaufmann Publishers Inc., San Francisco, CA,
  USA (2001)

\bibitem{NIPS2013_9aa42b31}
Mikolov, T., Sutskever, I., Chen, K., Corrado, G.S., Dean, J.: Distributed
  representations of words and phrases and their compositionality. In: Burges,
  C., Bottou, L., Welling, M., Ghahramani, Z., Weinberger, K. (eds.) Advances
  in Neural Information Processing Systems. vol.~26. Curran Associates, Inc.
  (2013),
  \url{https://proceedings.neurips.cc/paper/2013/file/9aa42b31882ec039965f3c4923ce901b-Paper.pdf}

\bibitem{mitkov2016computational}
Mitkov, R.: Computational phraseology light: automatic translation of multiword
  expressions without translation resources. Yearbook of Phraseology
  \textbf{7}(1),  149--166 (2016)

\bibitem{mitkov2018multiword}
Mitkov, R., Monti, J., Pastor, G.C., Seretan, V.: Multiword units in machine
  translation and translation technology, vol.~341. John Benjamins Publishing
  Company (2018)

\bibitem{moreau-etal-2018-crf}
Moreau, E., Alsulaimani, A., Maldonado, A., Vogel, C.: {CRF}-seq and
  {CRF}-{D}ep{T}ree at {PARSEME} shared task 2018: Detecting verbal {MWE}s
  using sequential and dependency-based approaches. In: Proceedings of the
  Joint Workshop on Linguistic Annotation, Multiword Expressions and
  Constructions ({LAW}-{MWE}-{C}x{G}-2018). pp. 241--247. Association for
  Computational Linguistics, Santa Fe, New Mexico, USA (Aug 2018),
  \url{https://aclanthology.org/W18-4926}

\bibitem{morgan-etal-2021-wlv}
Morgan, S., Ranasinghe, T., Zampieri, M.: {WLV}-{RIT} at {G}erm{E}val 2021:
  Multitask learning with transformers to detect toxic, engaging, and
  fact-claiming comments. In: Proceedings of the GermEval 2021 Shared Task on
  the Identification of Toxic, Engaging, and Fact-Claiming Comments. pp.
  32--38. Association for Computational Linguistics, Duesseldorf, Germany (Sep
  2021), \url{https://aclanthology.org/2021.germeval-1.5}

\bibitem{pathirana2018knowledge}
Pathirana, N., Seneviratne, S., Samarawickrama, R., Wolff, S., Chitraranjan,
  C., Thayasivam, U., Ranasinghe, T.: Knowledge building via optimally
  clustered word embedding with hierarchical clustering. In: 15th International
  Conference on Natural Language Processing. p.~69 (2018)

\bibitem{pathirana2019concept}
Pathirana, N., Seneviratne, S., Samarawickrama, R., Wolff, S., Chitraranjan,
  C., Thayasivam, U., Ranasinghe, T.: Concept discovery through information
  extraction in restaurant domain. Computaci{\'o}n y Sistemas  \textbf{23}(3),
  741--749 (2019)

\bibitem{pennington-etal-2014-glove}
Pennington, J., Socher, R., Manning, C.: {G}lo{V}e: Global vectors for word
  representation. In: Proceedings of the 2014 Conference on Empirical Methods
  in Natural Language Processing ({EMNLP}). pp. 1532--1543. Association for
  Computational Linguistics, Doha, Qatar (Oct 2014). \doi{10.3115/v1/D14-1162},
  \url{https://aclanthology.org/D14-1162}

\bibitem{damith2022DTWquranqa}
Premasiri, D., Ranasinghe, T., Zaghouani, W., Mitkov, R.: Dtw at qur'an qa
  2022: Utilising transfer learning with transformers for question answering in
  a low-resource domain. In: Proceedings of the 5th Workshop on Open-Source
  Arabic Corpora and Processing Tools (OSACT5). (2022)

\bibitem{ranasinghe-etal-2021-exploratory}
Ranasinghe, T., Orasan, C., Mitkov, R.: An exploratory analysis of multilingual
  word-level quality estimation with cross-lingual transformers. In:
  Proceedings of the 59th Annual Meeting of the Association for Computational
  Linguistics and the 11th International Joint Conference on Natural Language
  Processing (Volume 2: Short Papers). pp. 434--440. Association for
  Computational Linguistics, Online (Aug 2021).
  \doi{10.18653/v1/2021.acl-short.55},
  \url{https://aclanthology.org/2021.acl-short.55}

\bibitem{rohanian-etal-2019-bridging}
Rohanian, O., Taslimipoor, S., Kouchaki, S., Ha, L.A., Mitkov, R.: {B}ridging
  the gap: {A}ttending to discontinuity in identification of multiword
  expressions. In: Proceedings of the 2019 Conference of the North {A}merican
  Chapter of the Association for Computational Linguistics: Human Language
  Technologies, Volume 1 (Long and Short Papers). pp. 2692--2698. Association
  for Computational Linguistics, Minneapolis, Minnesota (Jun 2019).
  \doi{10.18653/v1/N19-1275}, \url{https://aclanthology.org/N19-1275}

\bibitem{sarkar-etal-2021-fbert-neural}
Sarkar, D., Zampieri, M., Ranasinghe, T., Ororbia, A.: f{BERT}: A neural
  transformer for identifying offensive content. In: Findings of the
  Association for Computational Linguistics: EMNLP 2021. pp. 1792--1798.
  Association for Computational Linguistics, Punta Cana, Dominican Republic
  (Nov 2021). \doi{10.18653/v1/2021.findings-emnlp.154},
  \url{https://aclanthology.org/2021.findings-emnlp.154}

\bibitem{savary-etal-2017-parseme}
Savary, A., Ramisch, C., Cordeiro, S., Sangati, F., Vincze, V., QasemiZadeh,
  B., Candito, M., Cap, F., Giouli, V., Stoyanova, I., Doucet, A.: The
  {PARSEME} shared task on automatic identification of verbal multiword
  expressions. In: Proceedings of the 13th Workshop on Multiword Expressions
  ({MWE} 2017). pp. 31--47. Association for Computational Linguistics,
  Valencia, Spain (Apr 2017). \doi{10.18653/v1/W17-1704},
  \url{https://aclanthology.org/W17-1704}

\bibitem{schneider-etal-2016-semeval}
Schneider, N., Hovy, D., Johannsen, A., Carpuat, M.: {S}em{E}val-2016 task 10:
  Detecting minimal semantic units and their meanings ({D}i{MSUM}). In:
  Proceedings of the 10th International Workshop on Semantic Evaluation
  ({S}em{E}val-2016). pp. 546--559. Association for Computational Linguistics,
  San Diego, California (Jun 2016). \doi{10.18653/v1/S16-1084},
  \url{https://aclanthology.org/S16-1084}

\bibitem{taslimipoor2015using}
Taslimipoor, S., Mitkov, R., Pastor, G.C.: Using cross-lingual contexts to
  extract translation equivalents for multiword expressions from parallel
  corpora. In: Nuevos horizontes en los Estudios de Traducci{\'o}n e
  Interpretaci{\'o}n (Comunicaciones completas): Conferencia AIETI7 (29 al 31
  de enero de 2015 en M{\'a}laga). pp. 174--180. Editions Tradulex (2015)

\bibitem{wolf-etal-2020-transformers}
Wolf, T., Debut, L., Sanh, V., Chaumond, J., Delangue, C., Moi, A., Cistac, P.,
  Rault, T., Louf, R., Funtowicz, M., Davison, J., Shleifer, S., von Platen,
  P., Ma, C., Jernite, Y., Plu, J., Xu, C., Le~Scao, T., Gugger, S., Drame, M.,
  Lhoest, Q., Rush, A.: Transformers: State-of-the-art natural language
  processing. In: Proceedings of the 2020 Conference on Empirical Methods in
  Natural Language Processing: System Demonstrations. pp. 38--45. Association
  for Computational Linguistics, Online (Oct 2020).
  \doi{10.18653/v1/2020.emnlp-demos.6},
  \url{https://aclanthology.org/2020.emnlp-demos.6}

\bibitem{NEURIPS2019_dc6a7e65}
Yang, Z., Dai, Z., Yang, Y., Carbonell, J., Salakhutdinov, R.R., Le, Q.V.:
  Xlnet: Generalized autoregressive pretraining for language understanding. In:
  Wallach, H., Larochelle, H., Beygelzimer, A., d\textquotesingle
  Alch\'{e}-Buc, F., Fox, E., Garnett, R. (eds.) Advances in Neural Information
  Processing Systems. vol.~32. Curran Associates, Inc. (2019),
  \url{https://proceedings.neurips.cc/paper/2019/file/dc6a7e655d7e5840e66733e9ee67cc69-Paper.pdf}

\bibitem{zhuang-etal-2021-robustly}
Zhuang, L., Wayne, L., Ya, S., Jun, Z.: A robustly optimized {BERT}
  pre-training approach with post-training. In: Proceedings of the 20th Chinese
  National Conference on Computational Linguistics. pp. 1218--1227. Chinese
  Information Processing Society of China, Huhhot, China (Aug 2021),
  \url{https://aclanthology.org/2021.ccl-1.108}

\end{thebibliography}
%




\end{document}